\documentclass[sigconf]{acmart}
\usepackage{booktabs} % For formal tables
\usepackage{balance}
\usepackage{amsmath}
\usepackage{natbib}
\usepackage{multirow}
\usepackage{enumitem}
\usepackage{epsfig}
\usepackage{subcaption}
\usepackage{hhline}
\usepackage[T1]{fontenc}
\renewcommand\footnotetextcopyrightpermission[1]{} % removes footnote with conference information in first column
\DeclareMathOperator*{\argmax}{argmax} 
\graphicspath{{eps/}}
\DeclareGraphicsExtensions{.ps,.eps,.epsi}
\begin{document}
\title{A Detailed Look At CNN-based Approaches In Facial Landmark Detection}
\author{Chih-Fan Hsu$^{12}$, Chia-Ching Lin$^{12}$, Ting-Yang Hung$^3$, Chin-Laung Lei$^2$, and Kuan-Ta Chen$^1$}
\affiliation{
\vspace{2mm}
$^1$Institute of Information Science, Academia Sinica\\
$^2$Department of Electrical Engineering, National Taiwan University\\
$^3$Halicio\v{g}lu Data Science Institute, University of California San Diego\\
}

\renewcommand{\shortauthors}{}
\renewcommand{\shorttitle}{CNN-based Approaches In Facial Landmark Detection}

\begin{abstract}
Facial landmark detection has been studied over decades. Numerous neural network (NN)-based approaches have been proposed for detecting landmarks, especially the convolutional neural network (CNN)-based approaches. In general, CNN-based approaches can be divided into regression and heatmap approaches. However, no research systematically studies the characteristics of different approaches. In this paper, we investigate both CNN-based approaches, generalize their advantages and disadvantages, and introduce a variation of the heatmap approach, a pixel-wise classification (PWC) model. To the best of our knowledge, using the PWC model to detect facial landmarks have not been comprehensively studied. We further design a hybrid loss function and a discrimination network for strengthening the landmarks' interrelationship implied in the PWC model to improve the detection accuracy without modifying the original model architecture. Six common facial landmark datasets, AFW, Helen, LFPW, 300-W, IBUG, and COFW are adopted to train or evaluate our model. A comprehensive evaluation is conducted and the result shows that the proposed model outperforms other models in all tested datasets.
\end{abstract}
\vspace{2mm}
%
% The code below should be generated b1y the tool at
% CCS →  Computing methodologies →  Computer graphics →  Graphics systems and interfaces
% We no longer use \terms command

\maketitle
\section{Introduction} \label{ito}
Facial landmarks are the fundamental components for various applications.
For instance, the landmarks can be used to understand facial expressions~\cite{DBLP:journals/corr/abs-1804-08348}, estimate head poses~\cite{Ranjan2019}, recognize faces~\cite{7243358}, or manipulate facial components~\cite{10.1007/978-3-319-46475-6_20}.
Facial landmark detection aims to automatically detect facial landmarks in an image and has been studied over decades~\cite{Wang2018}.

Various approaches were proposed to detect facial landmarks, such as active appearance models (AAM)~\cite{927467} and constrained local models (CLM)~\cite{CLM}.
Using deep learning approaches to detect landmarks gradually dominates this research topic.
Numerous neural network (NN)-based approaches are proposed for detecting landmarks, especially the convolutional neural network (CNN)-based approaches.
Generally, CNN-based approaches can be further divided into regression and heatmap approaches. Regression approaches directly infer the horizontal and vertical coordinates from a facial image; heatmap approaches detect the spatial position in a set of two-dimension heatmaps.
Many researchers dedicate to develop various network architectures for both approaches and prove that both approaches are robust and accurate for detecting landmarks.
However, there is no systematic research to study the characteristics of both approaches. 

Instead of developing a new network architecture for detecting facial landmarks, in this paper, we (1) investigate both CNN-based approaches, (2) generalize their advantages and disadvantages, and (3) investigate a variation of the heatmap approach, the pixel-wise classification (PWC) model.
Although, the PWC model is widely used for the object instance segmentation~\cite{Shelhamer2017,He2017a} and the joint detection~\cite{Bulat2016a},
to the best of our knowledge, using the PWC model to detect facial landmarks have not been comprehensively studied.
Besides, detecting facial landmarks by the PWC model might be problematic because numerous landmarks are located at positions with similar image structure.
To further improve the detection accuracy of the PWC model by integrating the advantages from the regression and heatmap approaches, we design a hybrid loss function and a discrimination network to strengthen the landmarks' interrelationship implied in the PWC model.

Six facial landmark datasets, AFW dataset~\cite{6248014}, Helen dataset~\cite{Le:2012:IFF:2403072.2403124}, Labeled Face Parts in the Wild (LFPW) dataset~\cite{6412675}, 300 Faces In-the-Wild Challenge (300-W) dataset, and the additional 135 images in difficult poses and expressions of the 300-W dataset (IBUG), and the testing set of the Caltech Occluded Faces in the Wild (COFW) dataset~\cite{Burgos-Artizzu_2013_ICCV}, are adopted to train or evaluate our model.
We conduct a comprehensive evaluation and the result shows that the detection accuracy of the PWC model can be improved without modifying the model architecture.
Besides, the proposed model outperforms other models in all datasets.
For making the community reproduce our experiment and develop further, we also release the source codes of all models on the GitHub\footnote{\url{https://github.com/chihfanhsu/fl_detection}}. 

%To summarize this paper, our contributions are listed as follows:
%\begin{itemize}[topsep=0pt, partopsep=0pt, itemsep=0.5pt, parsep=0.5pt, leftmargin=.25in]
%\item {we investigated the CNN-based regression and heatmap approaches and generalized their advantages and disadvantages;}
%\item {we investigate the pixel-wise classification model for detecting facial landmarks;}
%\item {we propose a hybrid loss function and a discrimination network to further improve the detection accuracy of the PWC model; and}
%\item {we conduct a comprehensive experiment for evaluating the proposed method on six common facial landmark datasets. The result shows that our model outperforms the other investigated models.}
%\end{itemize}
%\input{rel}
\section{Convolutional Neural Network-Based Approaches} \label{rel}
Using deep learning models to detect facial landmarks is a popular research topic because CNN-based approaches become more efficient and powerful. Generally, CNN-based approaches can be divided into regression and heatmap approaches.
\subsection{Regression Approaches} \label{reg}
The regression approaches can be further divided into direct and cascaded regression models.
%\subsubsection{Direct regression models}

\begin{figure}[t]
\centering
\resizebox{.2\textwidth}{!}{
    \centering{
        \includegraphics[width=0.6\textwidth]{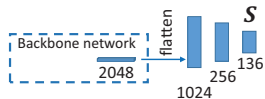}
    }
}
\caption{An example model architecture of the direct regression model for detecting 68 facial landmarks. A rectangle denotes a fully-connected layer and the number of nodes in the layer is listed beneath the rectangle.}
\label{fig:reg_structure}    
\end{figure}

\textbf{Direct regression models}.
Detecting facial landmarks by direct regression models is studied for many years~\cite{Bulat2016,Zhang2016,Yang2017,Zadeh2017,Ranjan2017,Wu2018a,Wu2018,Ranjan2019}.
The model detects the landmark coordinates $S$ represented by a vector from a facial image $I$.
The dimension of the vector is the twice number of landmarks.
Figure~\ref{fig:reg_structure} shows an example of a direct regression model for detecting 68 facial landmarks.
The backbone network can be any network architecture for extracting features from the input facial image. Here, we assume that the last layer of the backbone network contains 2,048 channels.
%
%\subsubsection{Cascaded regression models}

\textbf{Cascaded regression models}.
Unlike direct regression models that directly detect landmark coordinates, the cascaded regression models iteratively update a predefined or a pre-detected landmarks $S_0$ to detect landmarks~\cite{Sun2013,Zhou2013,Fan2016,Lv2017,Zhang2014a,
Kowalski2017,Zhang2016a,He2017,Chai2016,Dong2018}.
A sub-network is used to generate an updating vector $\Delta S_{i}$ to update the landmark positions in each stage $i$.
After $n$ updates, the model outputs the final landmark coordinates $S_n$.
Generally, the cascaded regression models can be simply formulated as $S_i = S_{i-1} + \Delta S_{i-1}, i={1,2,...,n}$. 

To train a regression approach, the L2 distance is adopted to evaluate the point-wise difference between the detected and the ground-truth landmarks, which can be formulated as
\begin{equation}
loss_{reg} = \dfrac{1}{|L|}\sum_{l\in L}\|S_l-\hat{S}_l\|_2,
\end{equation}
where $L$ is the set of landmarks, $S_l$ and $\hat{S}_l$ represent the detected and the ground-truth coordinates of the $l^{th}$ landmark, respectively.

Comparing the direct and the cascaded models, cascaded regression models generally are more effective than direct regression models because cascaded regression models follow the coarse-to-fine strategy~\cite{Wu2019}.
However, there is no standard to define how many stages should be involved in a cascaded model to achieve the best detection accuracy.
Also, there is no standard to generate the predefined face shape.
Hence, numerous studies obtained the predefined shape by averaging face shapes of the training set or predict the shape by an additional model. 

\begin{figure}[t]
\centering
\resizebox{.4\textwidth}{!}{
    \centering{
        \includegraphics[width=0.6\textwidth]{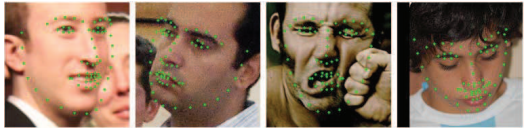}
    }
}
\caption{Detected results of the regression approach. The face shape formed by detected landmarks is maintained even detection failed. However, the landmarks suffer from inaccurate positions.}
\label{fig:res_reg}    
\end{figure}

Generally, regression approaches benefit from the strong interrelationship between landmarks because the structural information of a face is implicitly embedded and learned in the fully connected layers. Therefore, the detected landmarks can still form a face-like shape even if some key parts of a face are occluded. On the other hand, since the interrelationship is strong, the approach suffers from slightly inaccurate landmark positions. Namely, the approach tends to maintain the shape of landmarks as a face rather than detecting the true positions of landmarks.
Once detecting failed, the model may randomly place landmarks with a face-like shape. Figure~\ref{fig:res_reg} shows several examples detected by regression approaches.
The left two images illustrate examples that the detected landmarks suffer from inaccurate positions.
The right two images illustrate examples that the detection is failed but the shape of landmarks remains a face-like shape.
\subsection{Heatmap Approaches} \label{htm}
Heatmap approaches detect a landmark by indicating the position of the landmarks in a two-dimensional heatmap.
Generally, the model structures of heatmap approaches are inspired by the fully convolutional network (FCN) proposed by~\cite{Shelhamer2017}, which contains a convolutional part to generate semantic features from the facial image and a de-convolutional part to decode semantic features to a set of heatmaps.
According to the characteristic of the heatmap and the loss function, heatmap approaches can be divided into the distribution, the heatmap regression, and the pixel-wise classification models.

\begin{figure}[t]
\centering
\resizebox{.3\textwidth}{!}{
    \centering{
        \includegraphics[width=0.6\textwidth]{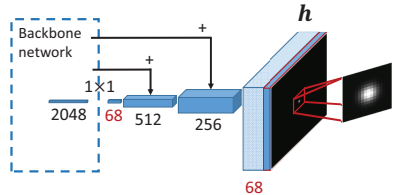}
    }
}
\caption{An example of the network architecture of the distribution model. The goal of the model is to predict a set of heatmaps as similar to the corresponding ground-truth maps generated by the multivariate distribution. Here, the distribution is the multivariate Gaussian with the three-pixel standard deviation. The cube denotes a tensor and the channel size is listed beneath the cube.}
\label{fig:dist_structure}
\end{figure}

\textbf{Distribution models}.
The distribution model indicates the position of a landmark by a multivariate distribution.
The center of the distribution is located at the landmark coordinates.
Generally, a two-dimensional Gaussian distribution is commonly used to indicate the landmark~\cite{Jackson2016, Bulat2017, Bulat2018, Robinson2019}.
Figure~\ref{fig:dist_structure} shows an example of the de-convolutional part of a distribution model for detecting 68 facial landmarks.
We mention here that the number of channels contained in the heatmaps is the same as the number of landmarks. A spatial softmax function is used to force the sum of elements in each heatmap equal to one.

To train a distribution model, Kullback-Leibler (KL) divergence is adopted to measure the distance between the predicted heatmap $h_l$ and the ground-truth Gaussian distribution $\hat{h}_l$ for each landmark $l$.
Namely, the loss can be calculated by $loss_{dist} = \sum_{l\in L}KL(\hat{h}_l \| h_l)$,
where $KL(\cdot)$ is the KL divergence.
%Mathematically, the KL divergence can be further simplified to the cross-entropy. Namely, the loss function is replaced by
%\begin{equation}
%loss_{distribution} = -\sum_{l\in L}\sum_{p\in I}\hat{h}_l (p)log(h_l(p)).
%\end{equation}
%

\textbf{Heatmap regression models}. 
Wei et al.~\cite{Wei2016} proposed an alternative model to detect landmarks from the heatmaps.
Although the proposed model, Convolutional Pose Machines, is used to detect landmarks of the human, the model can be extended to detect facial landmarks.
Differing from the distribution model, the output of the model contains an additional heatmap to indicate the background pixels.
The L2 loss is adopted to evaluate the difference between the ground-truth distribution maps and the predicted heatmaps.
Namely, the loss function can be formulated as $loss_{hreg} = \sum_{l\in \{L,b\}} (\hat{h}_l - h_l)^2$, where $b$ represents the image background.

\textbf{Pixel-wise classification models}.
Using pixel-wise classification (PWC) models to detect landmarks of the human are studied by He et al.~\cite{He2017a}.
The model also can be extended to detect facial landmarks, however, to the best of our knowledge, adopting PWC models to detect facial landmarks has not been comprehensively studied.

\begin{figure}[t]
\centering
\resizebox{.3\textwidth}{!}{
    \centering{
        \includegraphics[width=0.6\textwidth]{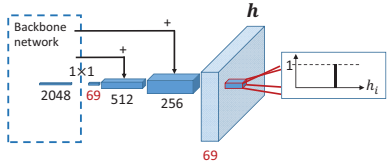}
    }
}
\caption{An example of the pixel-wise classification model. The vector at each pixel in output heatmaps is a probabilistic vector that indicates the pixel belonged to which landmark classes or the background class.}
\label{fig:pwc_structure}
\end{figure}

Despite the output of the PWC model is a set of heatmaps, which are similar to the aforementioned heatmap models, the meaning of the heatmap is different.
The heatmaps of the distribution and heatmap regression models are generated by the multivariate distribution to indicate the spatial position of the landmarks; the heatmap of the PWC model is generated by a set of probabilistic vectors that indicates the probabilities of the pixel belonged to which landmark or the background.
Specifically, each vector contains $|L|+1$ elements and the sum of elements is equal to one.
Figure~\ref{fig:pwc_structure} shows an example of the network architecture of the PWC model.

To train a PWC model, the cross-entropy loss is adopted.
Namely,
\begin{equation}
loss_{PWC} = -\dfrac{1}{|I|}\sum_{p\in I}\sum_{i\in \{L,b\} }\hat{h}_i^plog(h_i^p),
\end{equation}
where $h_i^p$ and $\hat{h}_i^p$ represent the predicted probabilistic vector and the ground-truth vector at pixel $p$, respectively, $i$ indicates the $i^{th}$ element of the vector. 

\begin{figure}[t]
\centering
\resizebox{.45\textwidth}{!}{
    \centering{
        \includegraphics[width=0.6\textwidth]{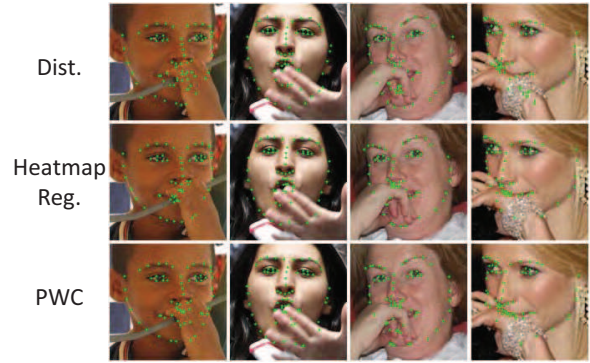}
    }
}
\caption{Several fail results detected by the heatmap approaches. The face shape suffers from serious distortion when some landmarks are not successfully detected. On the other hand, the successfully detected landmarks benefit from accurate positions.}
\label{fig:res_heatmap}
\end{figure}

To obtain the landmark positions from heatmaps, the coordinates of a landmark can be calculated by the position with the maximum probability in each heatmap, which can be calculated by $S_l = \argmax_{(x,y)}h_l, l\in L$.
The heatmap approaches benefit from highly accurate landmark positions but suffer from the lack of the interrelationship between landmarks.
It is because the interrelationship gradually degrades when it passes through de-convolutional layers.
Once some landmarks are detected failed (mostly caused by the occlusion), the detected position tends to shift to a nearby corner or edge.
As a result, the shape of the detected landmarks is distorted.
Figure~\ref{fig:res_heatmap} shows the detected results of the heatmap approaches.
As we can observe, landmarks corresponding to occluded parts are commonly missing and the face shape formed from the detected landmarks is distorted. 
\section{A Pixel-wise Classification Model with Discriminator}
We are seeking for a new method that contains the advantages and restrains from the disadvantages of regression and heatmap approaches.
Considering that the landmark accuracy is highly important, we develop a new model based on the heatmap approach.
Moreover, because the PWC model outperforms the other two heatmap models, we design a new model based on the PWC model (the accuracy will be discussed in Section~\ref{exp}). 

\textbf{A hybrid loss function}.
To overcome the disadvantages of heatmap approaches that the interrelationship among landmarks cannot be maintained, we introduce a hybrid loss function that combines the L2 loss and the PWC loss functions. The idea of the hybrid loss function is augmenting the PWC loss by the L2 distance by penalizing the landmark shifting when detecting failed to strengthen the interrelationship implied in the model. Our loss function can be formulated by
\begin{equation}
loss_{hybrid} = \alpha \times loss_{PWC} + \beta \times loss_{reg},
\end{equation}
where the hyperparameters $\alpha$ and $\beta$ are used to balance between loss functions and we empirically set $\alpha$ to 1 and $\beta$ to 0.25 in our experiment. 

\textbf{A discrimination network}.
We also expect that the detected landmarks should form a face-like shape.
To achieve this goal, we add a discrimination network $D$ after the detection network to encourage the detected landmarks to remain a face-like shape.
Specifically, the loss function is modified by $loss_{total} = loss_{hybrid} + loss_{face}$, where $loss_{face}$ is defined by $-E[log(D(S))]$, reflecting the encouragement of the predicted landmarks being classified to have a face-like shape given by the discrimination network.

The detection and discrimination networks are jointly trained as training a Generative Adversarial Networks proposed by Goodfellow et al.~\cite{Goodfellow2014}.
Specifically, in each training step, the training process can be divided into updating the detection network and updating the discrimination network stages.
The $loss_{total}$ is used to update the detection network and the loss function, $loss_{disc} = -(E[log(D(\hat{S}))] + E[log(1-D(S))])$, is used to update the discrimination network. In our experiment, in each training step, we empirically train the discrimination network once and the detection network twice.

\begin{figure}[t]
\centering
\resizebox{.35\textwidth}{!}{
    \centering{
        \includegraphics[width=0.6\textwidth]{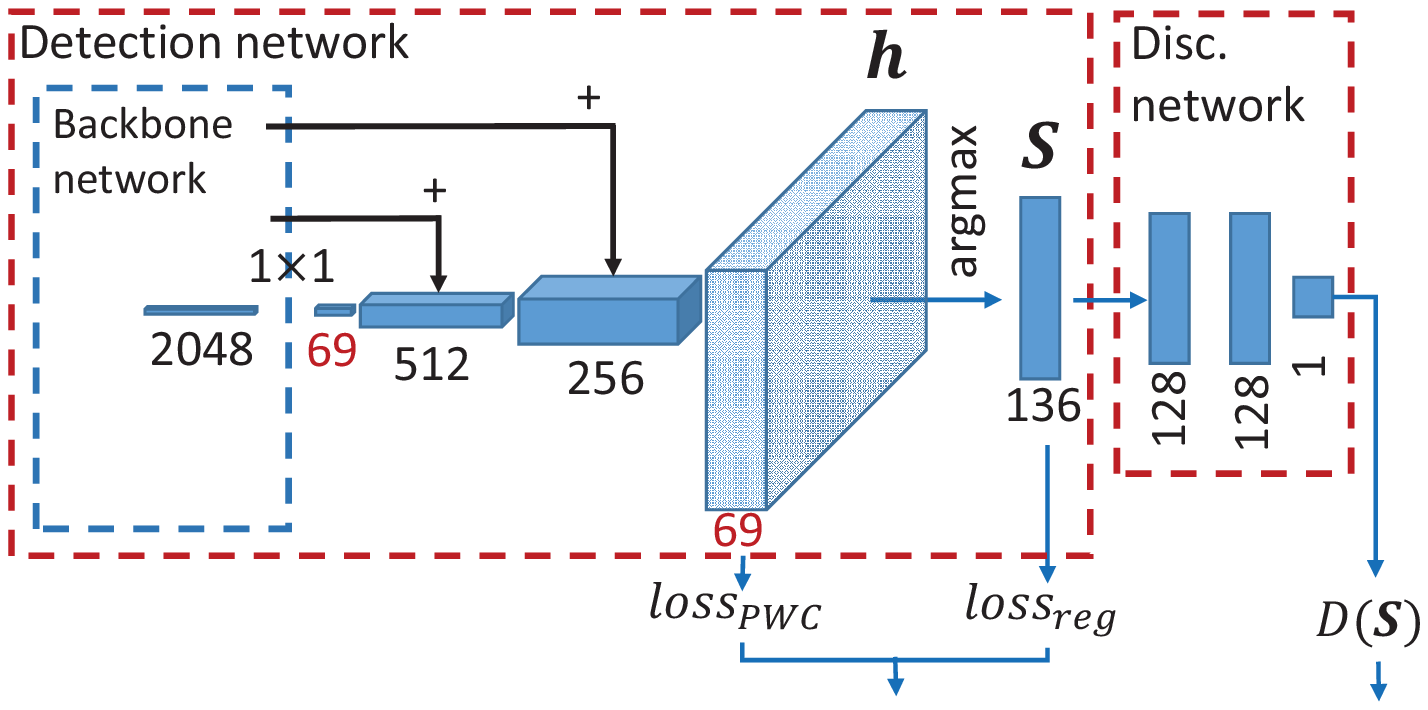}
    }
}
\caption{The proposed model contains a detection network and a discrimination network. The detection network is a PWC model for detecting landmarks. The discrimination network is a 3-layer fully connected NN for testing the shape of detected landmarks forms a face-like shape or not.}
\label{fig:mdl_structure}
\end{figure}

Figure~\ref{fig:mdl_structure} shows the network architecture of our model.
As we previously mentioned, the backbone network can be any network architecture.
A cube in the figure represents a specific tensor size and the number of channels of the tensor is shown below the cube. At least one convolutional block is performed under a certain tensor size. The convolutional block comprises a convolutional layer, a batch normalization layer, and an activation layer sequentially.
Once more than one convolutional block is performed, a $\times$ mark will be shown after the channel size and the number after the $\times$ mark indicates the number of convolutional blocks (Figure~\ref{fig:bb_structure}). The number above the cube denotes the kernel size of filters in the convolutional layer. The default kernel size is set to $3\times 3$ and omitted in the figure. The rectangular denotes a fully connected block and the number below each is the number of nodes in the fully connected layer. A fully connected block comprises a fully connected layer, a batch normalization layer, and an activation layer sequentially. 
\section{Training and Implementation} \label{imp}
\begin{figure}[t]
\centering
\resizebox{.30\textwidth}{!}{
    \centering{
        \includegraphics[width=0.6\textwidth]{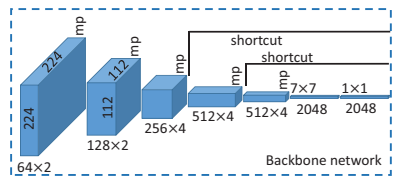}
    }
}
\caption{The backbone network. Each cube represents a tensor with a specific size. We only show the first two tensors for figure conciseness. Beneath the cube, the number before and after the $\times$ mark indicate the tensor's channel size and the number of convolutional blocks performed in the tensor, respectively. The number of convolutional blocks will be omitted when only one convolutional block is performed. The mp represents the max-pooling layer with $2\times 2$ kernel size and stride two.}
\label{fig:bb_structure}
\end{figure}

We implement models based on Tensorflow 1.8.0 and Python 3.5.3. The input image size is set to $224\times 224\times 3$. The backbone network is implemented by the VGG19-like network architecture and two shortcut links are used to improve the feature utilization for the heatmap approaches (Figure~\ref{fig:bb_structure}). A data augmentation algorithm is adopted to generate various orientations and sizes of faces to increase the dataset diversity. Specifically, we randomly rotated the input image from $-30^\circ$ to $30^\circ$ and rescaled the image with the ratio from 0.6 to 1.0 before the input image being fed to the model. Adam optimizer is used to update the network parameters. An early stopping mechanism is adopted to prevent the model from overfitting. Specifically, we calculate the average validation loss once every 1,000 training steps and stop the training process when the loss does not decrease ten times in a row.
%
%\subsection{Training and validation sets} \label{tav}

Several public facial landmark datasets, AFW, Helen, LFPW, 300-W, IBUG, and COFW datasets, are trained or tested in our experiment.
We use the default settings of datasets to separate the training and validation sets for making a fair comparison.
Table~\ref{tab:tab_imgs} shows the number of images for both sets in each dataset. To train the model, we crop the faces for every facial image according to the ground-truth landmarks to reduce negative impacts from unstable face detection.
Specifically, we calculate the maximum distance of the horizontal and the vertical distances ($d_h$ and $d_v$) of landmarks for each image.
Then, the facial image is cropped by a square bounding box with $1.3\times max(d_h,d_v)$ side length and the center of the bounding box is located at the centroid of landmarks. Finally, the cropped image is resized to $224\times 224\times3$ image size.
\begin{table}[t]
\centering
\caption{The number of images in the training (T) and validation (V) sets.}
\label{tab:tab_imgs}
\begin{tabular}{rcccccc}
\hline
\multicolumn{1}{l}{} & AFW & Helen & LFPW & \begin{tabular}[c]{@{}c@{}}300-W \\ (In/Out)\end{tabular} & IBUG & COFW \\ \hline
\#T & 337 & 2,000 & 811 & 0/0 & 0 & 0 \\ \hline
\#V & 0 & 330 & 224 & 300/300 & 135 & 507 \\ \hline
\end{tabular}
\end{table}
\section{Experimental Resutls} \label{exp}
\begin{figure}[t]
\centering
\resizebox{.45\textwidth}{!}{
    \centering{
        \includegraphics[width=0.6\textwidth]{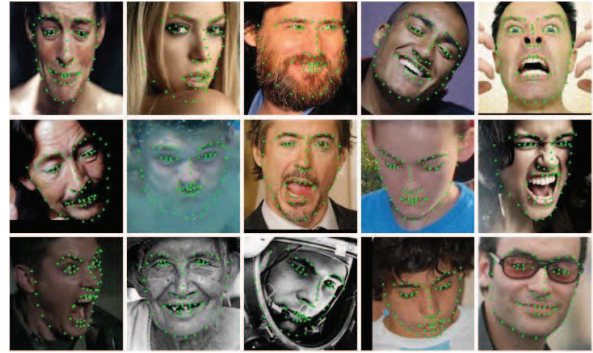}
    }
}
\caption{Several results detected by the hybrid+disc model.}
\label{fig:fig_res}    
\end{figure}

We name the model hybrid+disc for avoiding model garble in the following sections, where hybrid+disc denotes the model trained with $loss_{hybrid}$ and supported with the discrimination network during training.
Figure~\ref{fig:fig_res} shows the detected results of the hybrid+disc model.
As we can observe that the hybrid+disc model can handle various head orientations and facial expressions.
Also, the model can moderately handle partial occlusion. 

We adopt the normalized mean squared error (NMSE) to quantitatively evaluate the models. The NMSE can be calculated by $NMSE = \dfrac{1}{|L|}\sum_{l=1}^{|L|}\dfrac{\|S_l - \hat{S_l}\|_{2}}{d_{iod}}$, where $\hat{S_l}$ and $S_l$ represent the coordinates of the $l^{th}$ ground-truth and detected landmarks, respectively. The notation $d_{iod}$ represents the inter-ocular distance that can be calculated by the L2 distance between the outer corners of the ground-truth eyes. 

Nine models are compared in our experiment, (1) the Dlib model, an ensemble of regression trees model presented by Kazemi and Sullivan~\cite{Kazemi2014}, (2) the TCDCN model, a multi-task NN-based model presented by Zhang et al.~\cite{Zhang2016}, (3) a simple directly regression method implemented by CNN, (4) a cascaded regression model implemented by DAN model architecture, (5) a distribution model, the ground-truth heatmaps are generated by the Gaussian distribution with the three-pixels standard deviation, (6) a heatmap regression model, the ground-truth heatmaps are same as the distribution model, (7) a PWC model, a CNN-based model trained with $loss_{PWC}$, (8) a hybrid model, a CNN-based model trained with $loss_{hybrid}$, and (9) a hybrid+disc model.

To conduct a fair comparison among models, the backbone network in the investigated models is the same except the DAN model.
It is because the cascaded model contains several sub-networks.
Adopting the backbone network to all sub-networks leads the model to include massive parameters that exceed the memory limitation of our training machine.
Hence, we adopt the original three-stage DAN suggested by Kowalski et al.~\cite{Kowalski2017}.
Since the DAN model requires $112\times 112\times 3$ input image size, we resize the image size to meet the requirement and detect the landmark coordinates in $224\times 224$ image size.

In the experiment, the direct and cascaded regression models contain 178,682,440 and 35,053,144 parameters, respectively. The distribution model contains 82,871,880 parameters. 
The number of parameters of the PWC model and the heatmap regression model is the same because the difference between the two approaches is only the loss function, which contains 82,947,658 parameters.
The hybrid+disc model contains 82,982,347 parameters, in which the detection and discrimination networks contain 82,947,658 and 34,689 parameters, respectively.
Overall, the direct regression model contains the largest number of parameters.

Table~\ref{tab:tab_all} shows the average NMSE values of the nine models. We also evaluate the standard deviation from five independent trainings.
Since the standard deviations are small, we omit the standard deviation to make the table concise. 
The Dlib and the TCDCN models are the baselines for comparison.
The lowest NMSE value in each testing dataset is highlighted.
We mention there that the Dlib model cannot successfully detect the landmarks for all validation images.
Therefore, the average NMSE values of the Dlib model are only used for reference and the detection rates are listed below the table.

Generally, all investigated models outperform the base-line models except the heatmap regression model.
It is because the heatmap regression model suffers from the weak interrelationship between landmarks due to the network architecture and the small penalty for failure detection due to the loss function.
Therefore, once the landmark is detected failed, the detected position usually locates at the position that has a similar structure as the landmarks without occlusion and the position might be far from the ground-truth position.
The largest NMSE value in the IBUG dataset and the relative small NMSE value in the Helen dataset reveal this trend.
In between the regression models, surprisingly, the DAN model performs worse than the direct regression model.
We suspect that the DAN model is limited by the network architecture of the sub-network because the sub-network only contains 11,146,312 parameters that are much smaller than the number of parameters contained in the direct regression model.
In the heatmap models, the PWC model detects the landmarks more accurately than regression models and other heatmap models.
The hybrid model outperforms other models in almost all datasets.

\begin{figure}[t]
\centering
\begin{subfigure}[b]{0.22\textwidth}
\centering
\includegraphics[width=\textwidth]{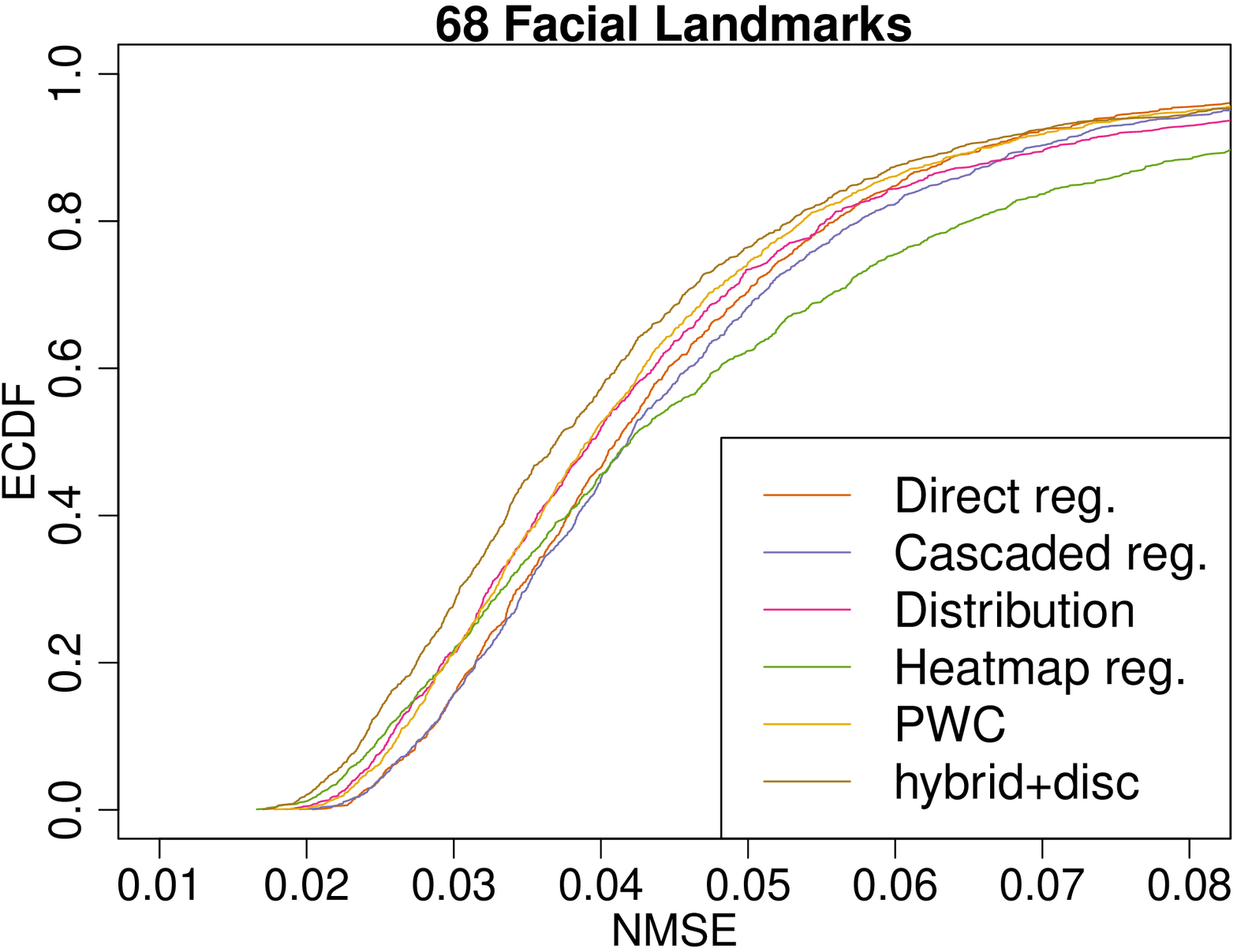}
\caption{68 facail landmarks\label{fig:all_loss}}
\end{subfigure}
\hfill
\begin{subfigure}[b]{0.22\textwidth}
\centering
\includegraphics[width=\textwidth]{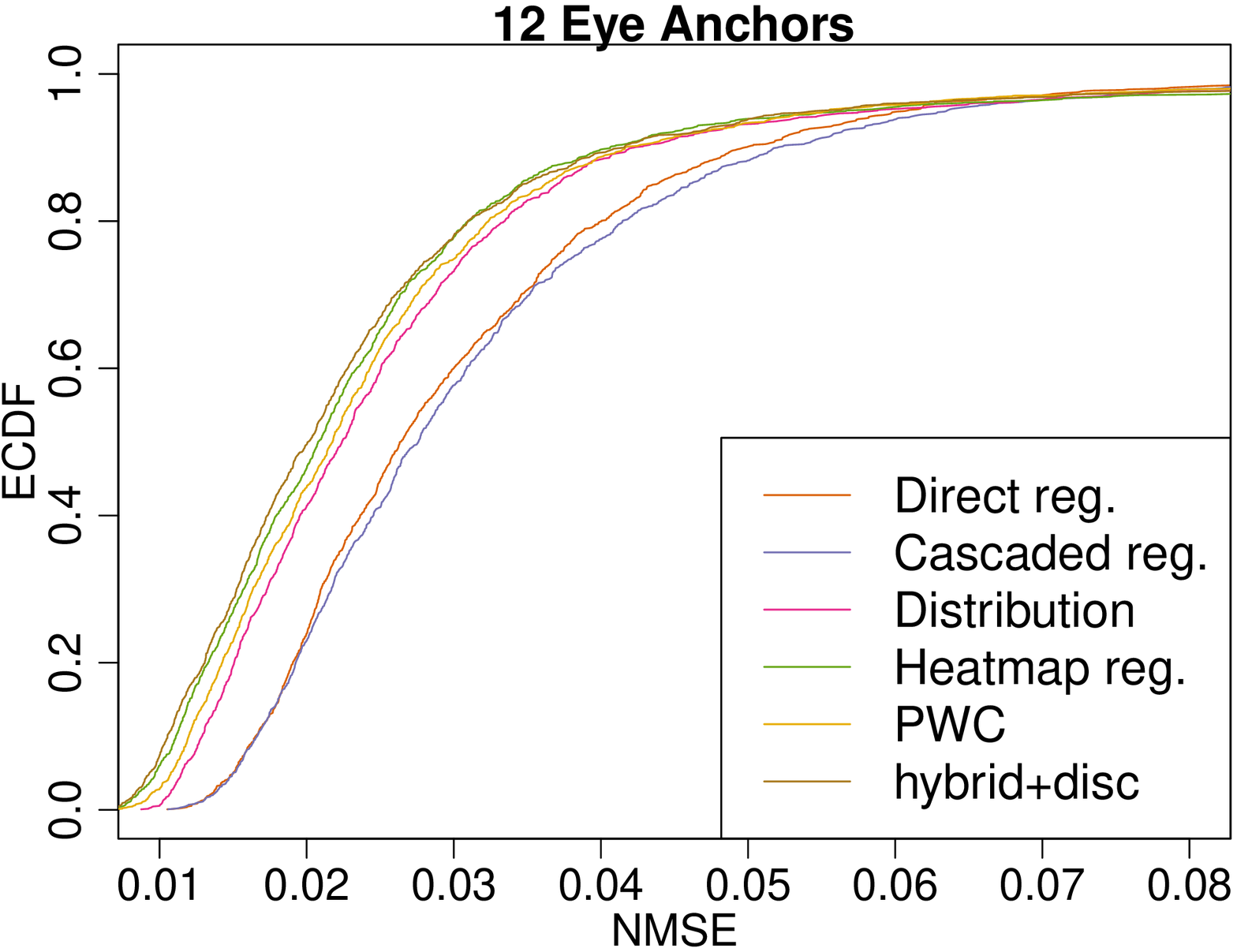}
\caption{12 eye anchors\label{fig:all_loss_eye}}
\end{subfigure}
\caption{The ECDF of the NMSE for the investigated models.}
\label{fig:eval_all}
\end{figure}

Figure~\ref{fig:eval_all} shows the empirical cumulative distribution function (ECDF) of NMSE values to illustrate the model effectiveness on the validation set.
Because the landmarks around the eyes (or called eye anchors) are more widely used than other landmarks, such as eye blink detection and gaze manipulation, we further evaluated the detection accuracy for the 12 eye anchors.
Figure~\ref{fig:all_loss} and Figure~\ref{fig:all_loss_eye} show the ECDF results of all 68 landmarks and 12 eye anchors, respectively.
As we previously mentioned that heatmap approaches generally achieve higher detection accuracy than the regression approaches.
However, the heatmap regression model suffers from large NMSE value when the landmarks are hard to detect. 
The PWC model slightly outperforms the distribution model.
The proposed hybrid loss function further improves the PWC model without enlarging the network architecture.
Overall, the hybrid model outperforms other investigated models. 
%
%\begin{table}[t]
%\centering
%\caption{The total number of parameters in the investigated models.}
%\label{tab:tab_paras}
%\begin{tabular}{lcr}
%\hline
%\multicolumn{1}{c}{Model} & Approach   & \#parameters \\ \hline
%Direct                    & Regression & 178,682,440   \\ \hline
%Cascaded (3-stage DAN)    & Regression & 35,053,144   \\ \hline
%Distribution (Gaussian)   & Heatmap    & 82,871,880   \\ \hline
%PWC                       & Heatmap    & 82,947,658   \\ \hline
%PWC+disc                  & Heatmap    & 82,982,347   \\ \hline
%hybrid                    & Heatmap    & 82,947,658   \\ \hline
%hybrid+disc               & Heatmap    & 82,982,347   \\ \hline
%\end{tabular}
%*The discrimination network contains 34,689 parameters.
%\end{table}
%
\begin{table*}[t]
\centering
\caption{The average NMSE of the investigated models for detecting 68 facial landmarks. The colored numbers indicate the smallest NMSE in each dataset. The 300-W dataset contains the indoor (In) and the outdoor (Out) subsets and the results are listed in two horizontal rows.}
\begin{tabular}{ccccccccccc}
\hline
Dataset & Dlib* & TCDCN & Direct & Cascaded & Dist. & Heat. reg. & PWC & PWC+disc & hybrid & hybrid+disc \\ \hline
Helen & 0.0298 & 0.0474 & 0.0356 & 0.0367 & 0.0337 & 0.0353 & 0.0334 & 0.0336 & 0.0315 & {\color[HTML]{FE0000} 0.0312} \\ \hline
LFPW & 0.0356 & 0.0454 & 0.0359 & 0.0360 & 0.0347 & 0.0371 & 0.0356 & 0.0359 & {\color[HTML]{FE0000} 0.0336} & 0.0337 \\ \hline
\begin{tabular}[c]{@{}c@{}}300-W\\ In/Out\end{tabular} & \begin{tabular}[c]{@{}c@{}}0.0677\\ 0.0640\end{tabular} & \begin{tabular}[c]{@{}c@{}}0.0781\\ 0.0746\end{tabular} & \begin{tabular}[c]{@{}c@{}}0.0495\\ 0.0500\end{tabular} & \begin{tabular}[c]{@{}c@{}}0.0520\\ 0.0516\end{tabular} & \begin{tabular}[c]{@{}c@{}}0.0508\\ 0.0503\end{tabular} & \begin{tabular}[c]{@{}c@{}}0.0597\\ 0.0586\end{tabular} & \begin{tabular}[c]{@{}c@{}}0.0499\\ 0.0487\end{tabular} & \begin{tabular}[c]{@{}c@{}}0.0505\\ 0.0492\end{tabular} & {\color[HTML]{FE0000} \begin{tabular}[c]{@{}c@{}}0.0481\\ 0.0470\end{tabular}} & \begin{tabular}[c]{@{}c@{}}0.0486\\ 0.0473\end{tabular} \\ \hline
IBUG & 0.0594 & 0.0795 & 0.0643 & 0.0678 & 0.0700 & 0.0812 & 0.0656 & 0.0663 & {\color[HTML]{FE0000} 0.0637} & 0.0639 \\ \hline
Total & 0.0496 & 0.0639 & 0.0452 & 0.0469 & 0.0455 & 0.0515 & 0.0446 & 0.0450 & {\color[HTML]{FE0000} 0.0427} & 0.0428 \\ \hline
\end{tabular} \label{tab:tab_all}

*Dlib does not successfully detect landmarks for all images in the validation set, the detection rate of Dlib in each dataset are listed as follows: Helen (318/330), LFPW (218/224), 300-W indoor (264/300) and outdoor (252/300), and IBUG (98/135).
\end{table*}
%
%\subsubsection{Occlusion tolerance} \label{ols}

\subsection{Ablation Study}
To ensure the hybrid loss function and the discrimination network can indeed improve the detection accuracy of the PWC model, we trained the models with several model and loss function combinations, which includes (1) the PWC model, (2) the PWC model supported by the discrimination network (PWC+disc), the hybrid model, and the hybrid+disc model.

Table \ref{tab:tab_all} shows the average NMSE of the investigated combinations.
Generally, the model trained with the hybrid loss function outperforms the model trained with the PWC loss function.
The result indicates that the regression loss function can support the PWC model to achieve better detection accuracy without modifying the architecture of the detection network.
However, the discrimination network does not significantly improve detection accuracy.

\begin{figure}[t]
\centering
\begin{subfigure}[b]{0.22\textwidth}
\centering
\includegraphics[width=\textwidth]{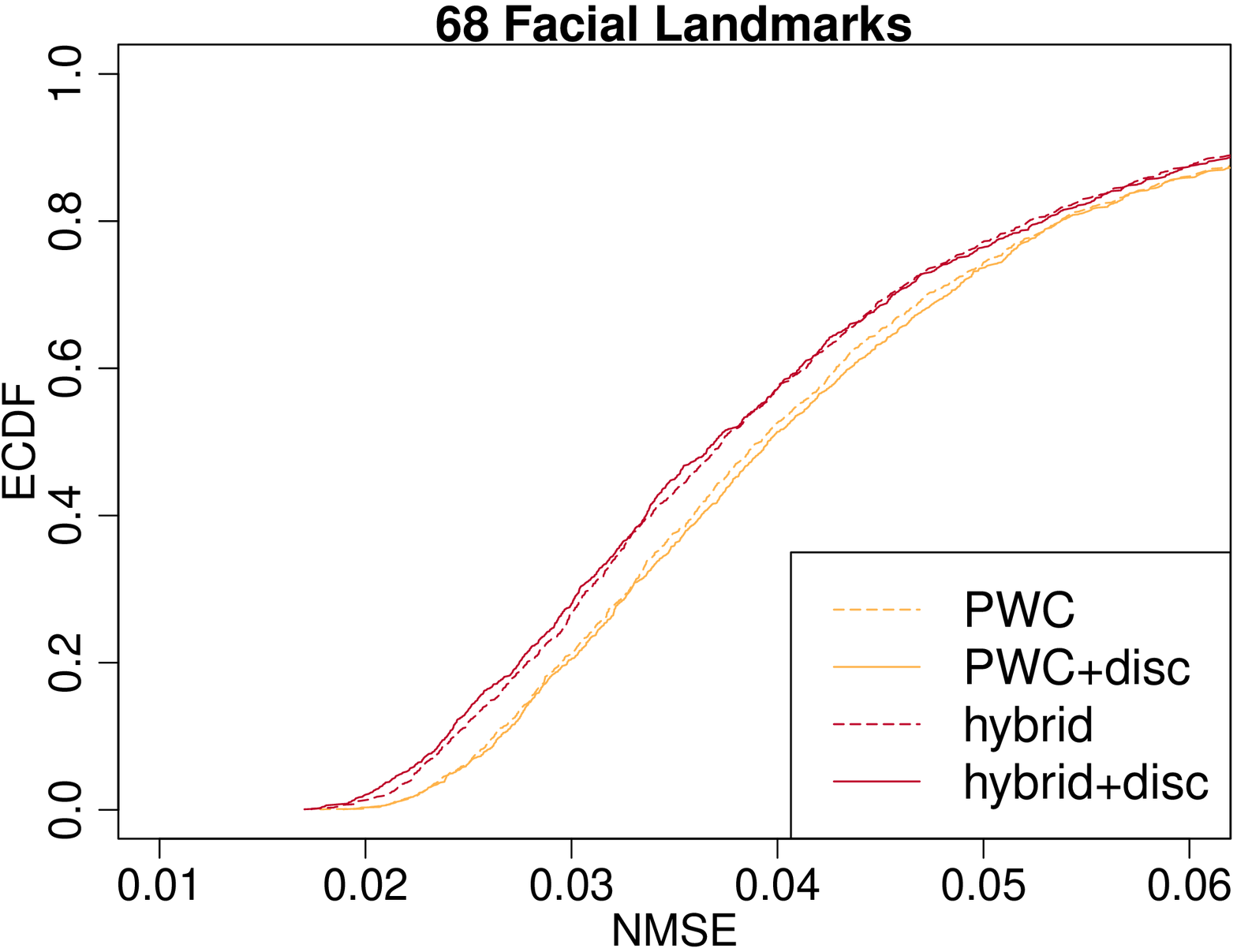}
\caption{68 landmarks\label{fig:abl_loss}}
\end{subfigure}
\hfill
\begin{subfigure}[b]{0.22\textwidth}
\centering
\includegraphics[width=\textwidth]{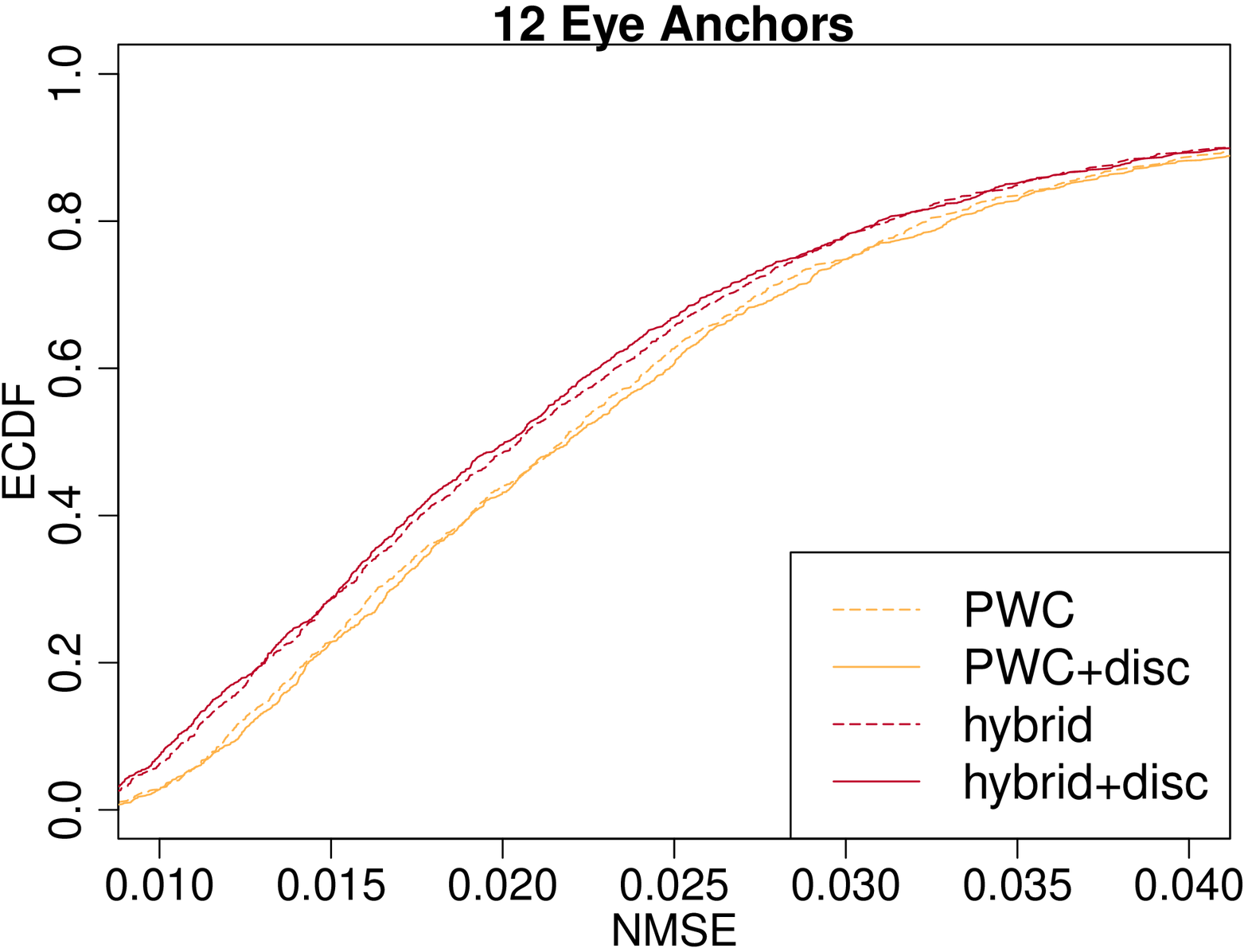}
\caption{12 eye anchors\label{fig:abl_loss_eye}}
\end{subfigure}
\caption{The ECDF of the NMSE values for different loss function combinations.}
\label{fig:eval_abl}
\end{figure}

Figure~\ref{fig:eval_abl} shows the ECDF of NMSE values for every loss combinations.
As we can observe that the hybrid model supported by the discrimination network achieves slightly better results than the models without supported by the discrimination network about $60\%$ and $80\%$ of validation images for detecting 68 facial landmarks and 12 eye anchors, respectively. However, for the PWC model, the discrimination network does not improve the detection accuracy.

\begin{figure}[t]
\resizebox{.48\textwidth}{!}{
    \centering{
        \includegraphics[width=0.6\textwidth]{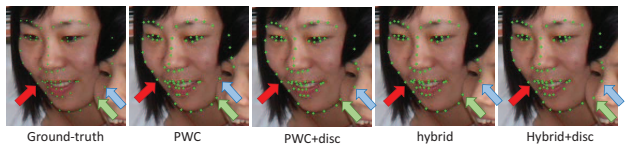}
    }
}
\caption{The hybrid loss function penalizes the landmark shifting due to the shape of the key point or partial occlusion. The discrimination network encourages the detected landmarks to form a face-like shape.}
\label{fig:n_138_move}
\end{figure}

To further explore the benefits by adopting the hybrid loss function and the discrimination network, we carefully observe the detected landmarks among the detected results.
Generally, the landmarks detected by the PWC model prefer to locate at the key point (the edge or the corner).
The result is not surprising because the key point is easier to be preserved in the semantic features than the smooth area and the facial landmarks should locate at the key point if the landmarks are not occluded.
However, the landmark positions are easy to be affected by the shape of the key point especially the landmarks located at the edge.
Besides, once the landmarks are occluded, the detected positions tend to located at a key point near the position where the landmark should be.
Therefore, the shape of the detected landmarks might be distorted.
Figure~\ref{fig:n_138_move} shows the illustration.
In the result detected by the PWC model, the landmark positions are affected by the shape of the edge (red and blue) and the occluded landmark located at a nearby corner (blue).
Adding the L2 loss to the PWC model penalizes the position shifting due to the occlusion or the negative impact from the shape of the key point, which greatly improves the detection accuracy.
However, the shape of the detected landmarks might still be distorted.
Although adopting the discrimination network in the training process might make landmarks slight inaccurate, the network successfully improves the shape of the landmarks.
Overall, adding the discrimination network has only a minor impact on detection accuracy. Besides, the discrimination network does not increase the inference time when testing. Training a model with a discrimination network is worth being considered and explored further.

\subsection{Occlusion Tolerance}
\begin{figure}[t]
\resizebox{.48\textwidth}{!}{
    \centering{
        \includegraphics[width=0.6\textwidth]{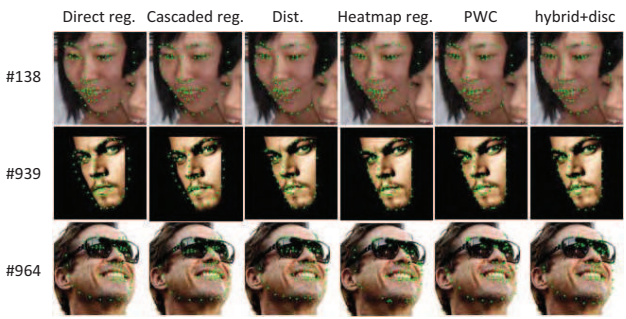}
    }
}
\caption{The hybrid+disc model can moderately guess the landmark positions when landmarks are occluded.}
\label{fig:fig_ocl}
\end{figure}

To explore the model's ability for handling facial images with partial occlusion, we mutually selected 298 images that some landmarks are occluded by objects or extreme light sources from the validation datasets.
Figure~\ref{fig:fig_ocl} shows several results detected by the investigated models.
As we can observe that the regression approaches suffer from inaccurate landmark positions.
The heatmap approaches benefit from the accurate positions but the landmarks shift due to the occlusion.
The hybrid+disc model eases the landmark shifting and can moderately guess the landmark positions when landmarks are occluded.

\begin{figure}[t]
\centering
\begin{subfigure}[b]{0.22\textwidth}
\centering
\includegraphics[width=\textwidth]{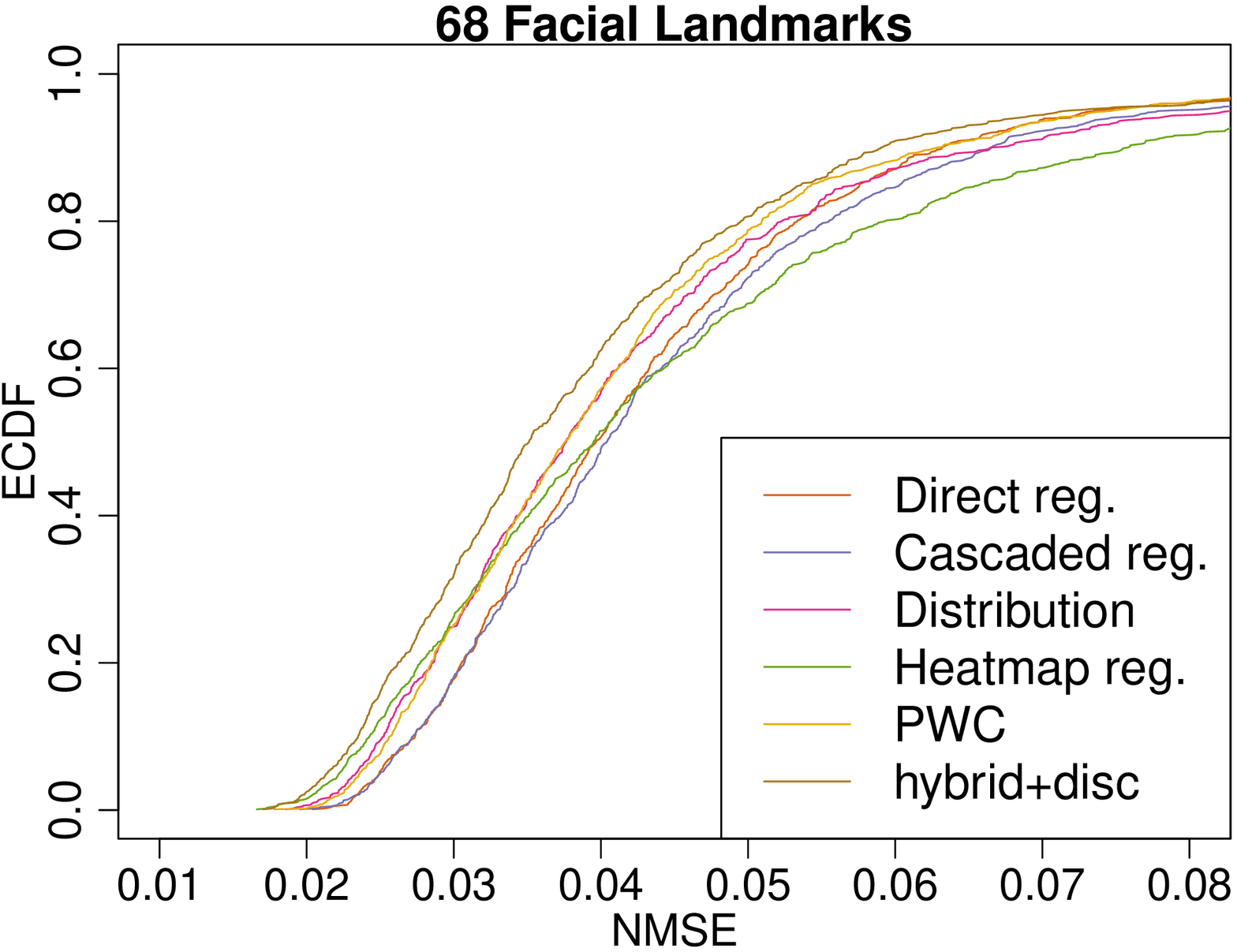}
\caption{Without occlusion\label{fig:without_ocd_loss}}
\end{subfigure}
\hfill
\begin{subfigure}[b]{0.22\textwidth}
\centering
\includegraphics[width=\textwidth]{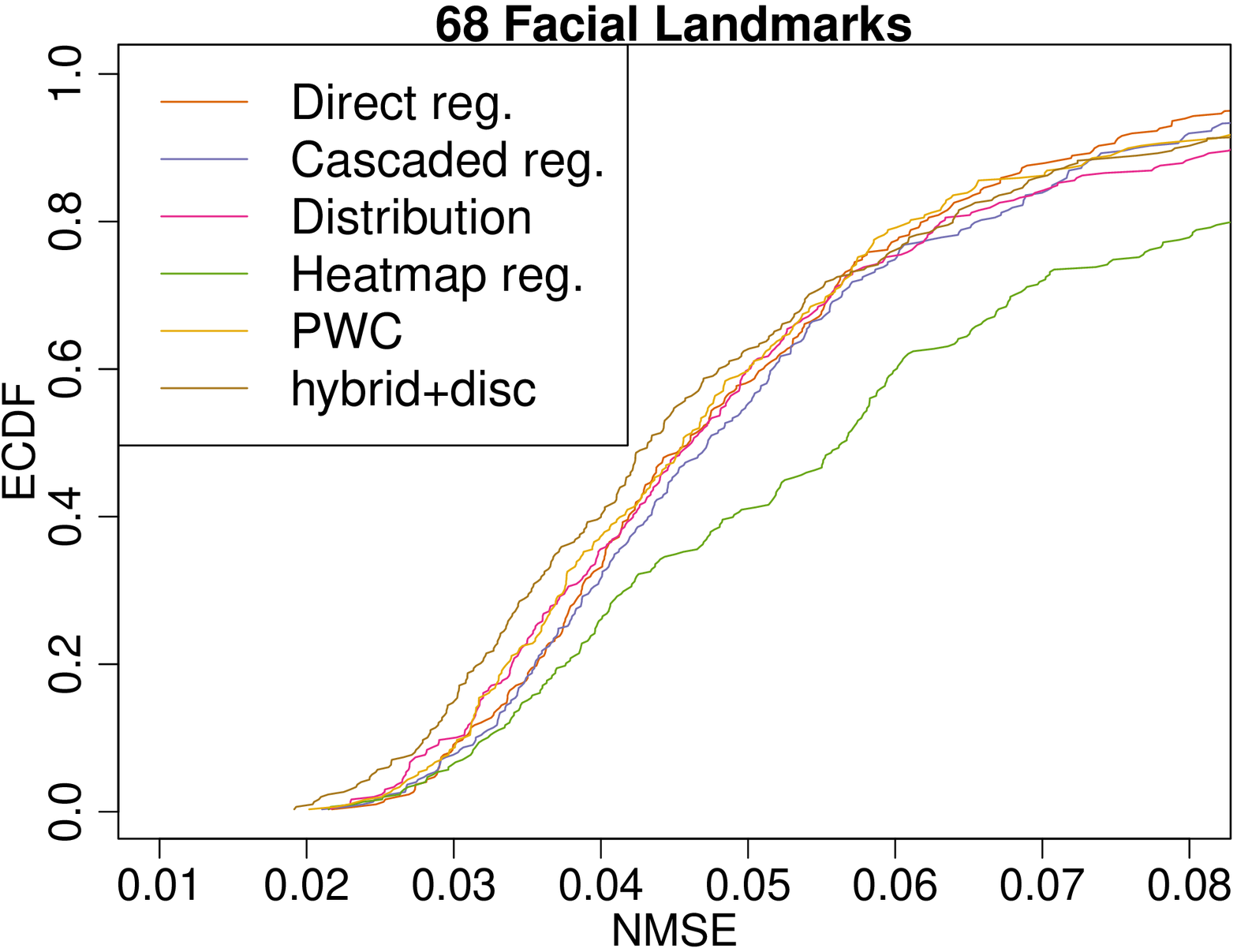}
\caption{With occlusion\label{fig:with_ocd_loss}}
\end{subfigure}
\caption{The ECDF of the NMSE values for evaluating occlusion tolerance.}
\label{fig:eval_occ}
\end{figure}

Figure~\ref{fig:eval_occ} shows the ECDF of the NMSE values for the images with and without partial occlusion.
As we can observe that the hybrid+disc model outperforms other models for the images without partial occlusion. For the images with partial occlusion, the hybrid+disc model achieves better model accuracy for about $70\%$ of validation images. It is worthwhile to mention that when the image has serious occlusion or failed detection, the regression approaches achieve lower NMSE values than heatmap approaches because the face shape is maintained.
%
%\subsubsection{Ablation study} \label{abl}

\begin{figure}[t]
\centering
\begin{subfigure}[b]{0.22\textwidth}
\centering
\includegraphics[width=\textwidth]{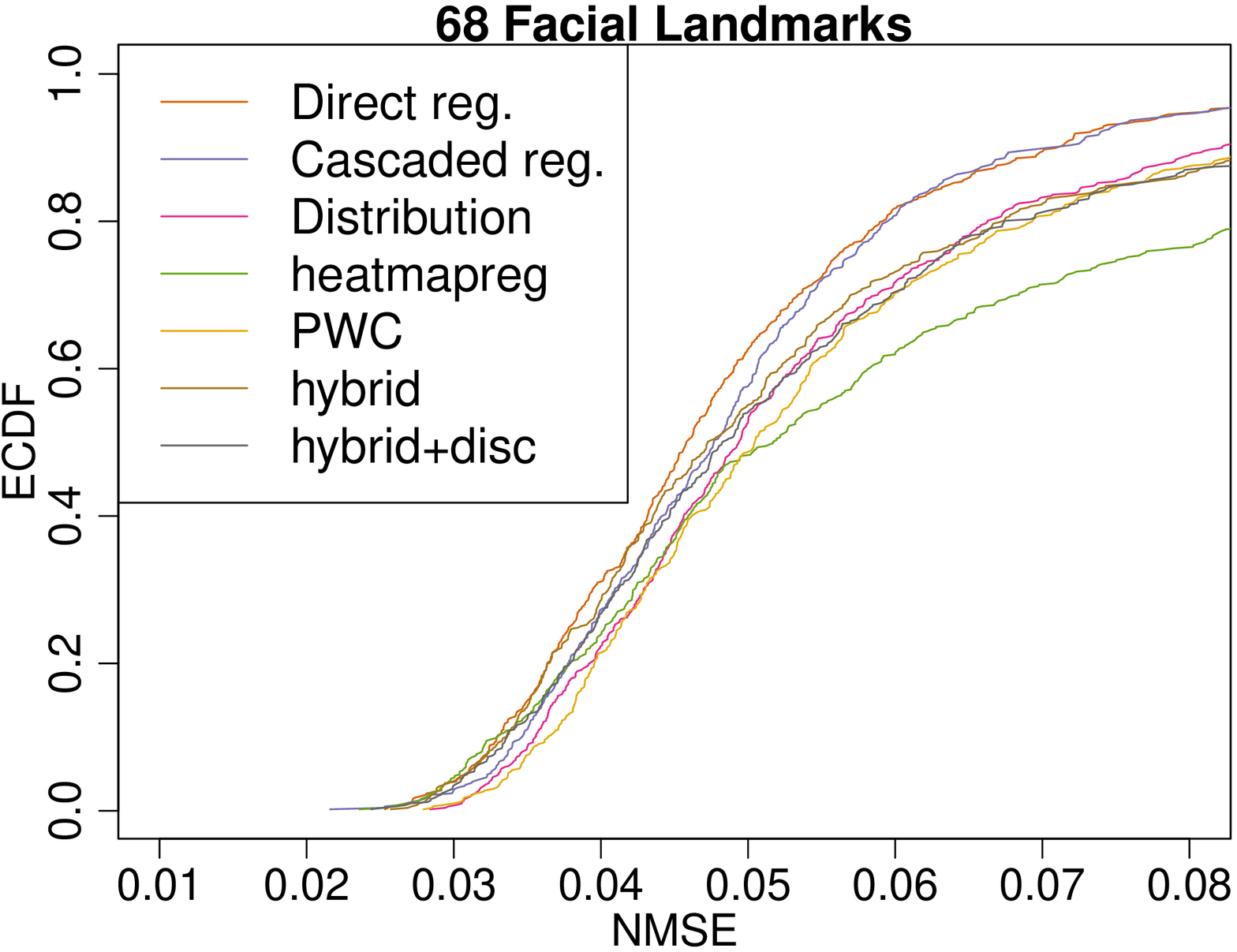}
\caption{68 facial landmarks\label{fig:cofw_loss}}
\end{subfigure}
\hfill
\begin{subfigure}[b]{0.22\textwidth}
\centering
\includegraphics[width=\textwidth]{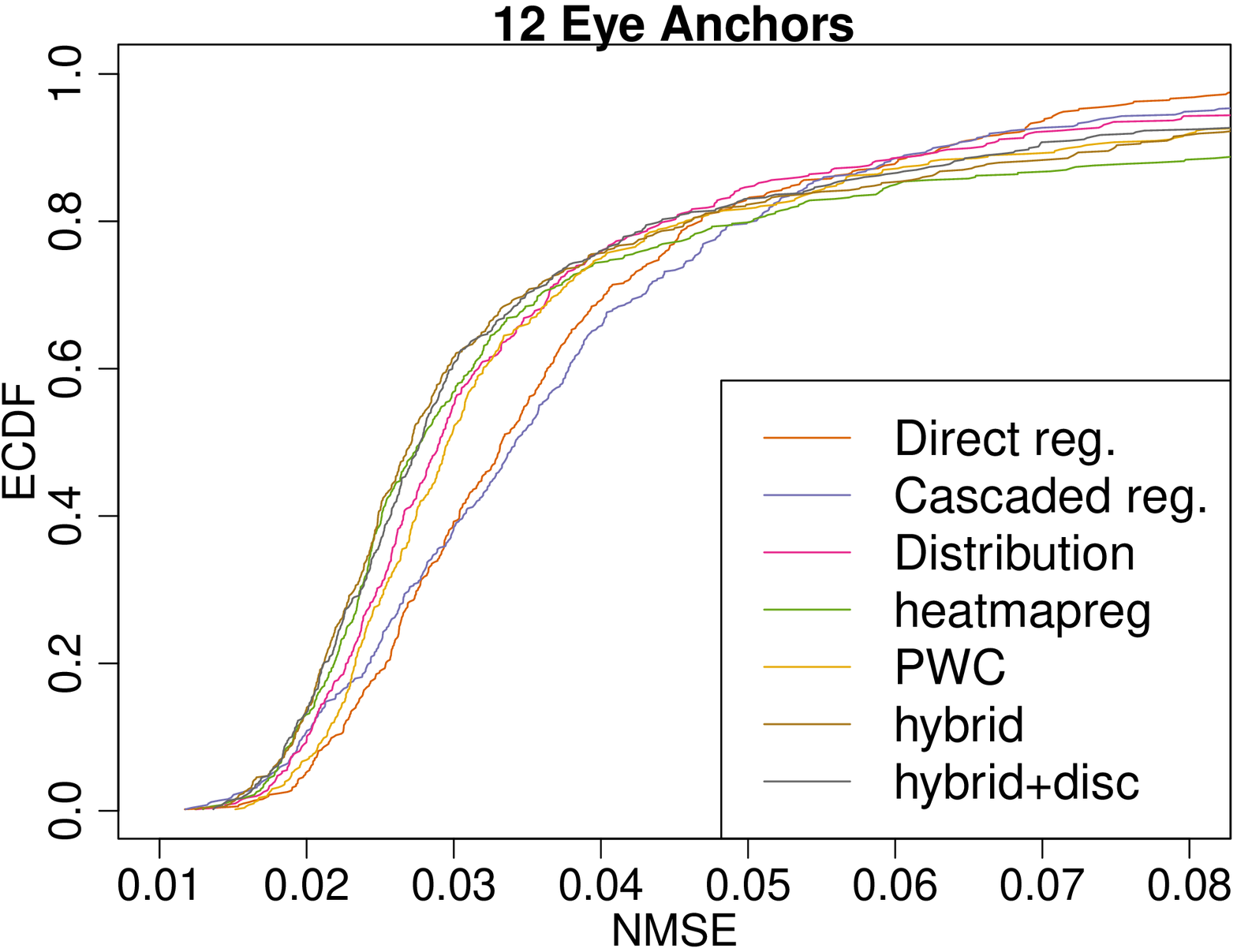}
\caption{12 eye anchors\label{fig:cofw_loss_eye}}
\end{subfigure}
\caption{The ECDF of the NMSE values for evaluating occlusion tolerance on the serious occluded images (the COFW dataset).}
\label{fig:eval_occ_cofw}
\end{figure}

To further explore the occlusion tolerance for the serious occluded facial images, we tested the investigated models on the testing set of the COFW dataset.
The testing set contains 507 images.
Figure~\ref{fig:cofw_loss} and Figure~\ref{fig:cofw_loss}, shows the ECDF value for detecting 68 facial landmarks and 12 eye anchors, respectively.
Figure~\ref{fig:cofw_loss} also reveals that the regression approaches perform better than the heatmap approaches in the images with serious occlusion.
The hybrid model slightly outperforms the other heatmap approaches about 70\% of testing images.
The heatmap regression performs the worst in all investigated model as the result in the previous experiment. In terms of the eye anchors, which is relatively less occluded than other landmarks, the heatmap approaches achieve a more precise landmark position than the regression approaches. Once the landmarks become hard to detect, the regression approaches gradually achieve better accuracy than the heatmap approaches.
Overall, the hybrid model and the hybrid+disc model outperforms other models in the 80\% of testing images. 
\section{Discussion and Conclusion} \label{discussion}
We have comprehensively studied the commonly used convolutional neural network-based approaches for detecting 68 facial landmarks, the regression and heatmap approaches, and their variations. We generalize the advantages and the disadvantages of these approaches and investigate a new variation of the heatmap model, pixel-wise classification (PWC) model.

To further improve the detection accuracy of the PWC model, we propose the hybrid loss function and comprehensively verify the loss function that can improve the detection accuracy without modifying the architecture of the detection network.
Besides, a discrimination network is proposed to encourage the detected landmarks to form a face-like shape.
Although the discrimination network has little impact on detection accuracy, it can be further studied.
Our proposed model is evaluated on six common facial landmark datasets, AFW, Helen, LFPW, 300-W, IBUG, and COFW datasets. The evaluated results reveal that the PWC model combined with the hybrid loss function achieves higher landmark accuracy than other investigated approaches not only for the 68 facial landmarks but also for the 12 eye anchors.
\bibliographystyle{unsrt}%\small
\bibliography{refs}
%\newpage
%\appendix
%\appendix
%\input{appendix}
\end{document}